%% file: main.tex
\documentclass[10pt,twocolumn,letterpaper]{article}

\usepackage{cvpr}
\usepackage{times}
\usepackage{epsfig}
\usepackage{graphicx}
\usepackage{amsmath}
\usepackage{amssymb}
\usepackage{graphicx}  \graphicspath{{figures/}}
\usepackage{algorithm}% http://ctan.org/pkg/algorithms
\usepackage{algpseudocode}% http://ctan.org/pkg/algorithmicx
\usepackage[normalem]{ulem}
\usepackage{makecell}
\usepackage{rotating}
\usepackage{booktabs}
\usepackage{gensymb}
\usepackage{capt-of}
\usepackage{enumitem}
\usepackage[dvipsnames,svgnames,x11names]{xcolor}
\usepackage{booktabs}
\usepackage{caption}
\usepackage{subcaption}
\usepackage{array}
\usepackage{multirow}
\usepackage{media9}
\usepackage{pbox}
\usepackage{authblk}
\usepackage[symbol]{footmisc}
\usepackage{url}
\cvprfinalcopy % *** Uncomment this line for the final submission

\def\cvprPaperID{457} % *** Enter the CVPR Paper ID here

\newcommand{\setmode}[1]{\def\mode{#1}}
\setmode{draft} % final OR draft
\long\def\IGNORE#1{} \long\def\COMMENT#1{}
\def\authornote#1#2#3{{\textcolor{#2}{\textsl{\small#1:[*#3*]}}}}

\ifthenelse{\equal{\mode}{draft}} { %%%%% ADD YOUR TAG HERE and USE IT FOR NOTES!
    \newcommand{\mbnote}[1]{\authornote{MB}{Red}{#1}} % Mark
    \newcommand{\vjnote}[1]{\authornote{VJ}{Magenta}{#1}} % Varun
    \newcommand{\khnote}[1]{\authornote{KH}{Blue}{#1}} % Kihwan   
    \newcommand{\hlnote}[1]{\authornote{HL}{Orange}{#1}} % Hendrik
    \newcommand{\JK}[1]{\authornote{JK}{cyan}{#1}} % Jan
    } {}

\ifthenelse{\equal{\mode}{final}} {
    \newcommand{\mbnote}[1]{}
    \newcommand{\khnote}[1]{}
    \newcommand{\vjnote}[1]{}
    \newcommand{\hlnote}[1]{}
    \newcommand{\JK}[1]{}
    \typeout{************* MODE: Final}
    } {}

\newcommand{\mycaption}[2]{\caption{\textbf{#1.}\xspace#2}}

\input{macros}

\title{Two-shot Spatially-varying BRDF and Shape Estimation}
\author{\vspace{-1em}%
Mark Boss${}^{1*}$,\quad %\footnote{Work done during internship at NVIDIA}
Varun Jampani${}^2$,\quad %
Kihwan Kim${}^2$,\quad %
Hendrik P.A. Lensch${}^1$,\quad%
Jan Kautz${}^2$
\\
${}^1$University of T{\"{u}}bingen,\quad%
${}^2$NVIDIA
}

\begin{document}

\twocolumn[{
            \maketitle
            \vspace{-1.5em}
            \centerline{
                \input{figures/teaser/Teaser.tex}}
            \captionof{figure}
            {
                \textbf{Practical SVBRDF and shape estimation.} \xspace {Sample two-shot input and the corresponding estimates for SVBRDF (albedo, specularity, roughness) and shape (depth and normals). The novel re-render are animated and show a moving view and light. We recommend Adobe Acrobat or Okular for viewing. Samples are taken from~\cite{Yagiz2018}.
                    %\JK{could we embed a video with object/lighting moving a bit?}
                }
            }
            \label{fig:teaser}
            \vspace{2em}
        }]

%\maketitle
%\thispagestyle{empty}

%%%%%%%%% BODY TEXT

\footnotetext[1]{Work done during an internship at NVIDIA.}
\footnotetext{Dataset and Code available at: \url{markboss.me/publication/cvpr20-two-shot-brdf}}

\input{sec_abstract}

\input{sec_introduction}

\input{sec_related}

\input{sec_approach}

\input{sec_results}

\input{sec_conclusion}

\clearpage

% \appendix
% \section*{Supplementary}

% \input{supplementary_content.tex}

{\small
    \bibliographystyle{ieee}
    \bibliography{flashbrdf}
}

\end{document}

%% file: macros.tex
\newcolumntype{L}[1]{>{\raggedright\let\newline\\\arraybackslash\hspace{0pt}}m{#1}}
\newcolumntype{C}[1]{>{\centering\let\newline\\\arraybackslash\hspace{0pt}}m{#1}}
\newcolumntype{R}[1]{>{\raggedleft\let\newline\\\arraybackslash\hspace{0pt}}m{#1}}

\newcommand{\sect}[1]{Sec.~\ref{#1}}

\newcommand{\fig}[1]{Fig.~\ref{#1}}
\newcommand{\tbl}[1]{Table~\ref{#1}}

\newcommand{\ignore}[1]{}

\makeatletter
\DeclareRobustCommand\onedot{\futurelet\@let@token\@onedot}
\def\@onedot{\ifx\@let@token.\else.\null\fi\xspace}

\def\eg{\emph{e.g}\onedot}

\def\etal{\emph{et al}\onedot}

\makeatother

\definecolor{MyDarkBlue}{rgb}{0,0.08,1}
\definecolor{MyDarkGreen}{rgb}{0.02,0.6,0.02}
\definecolor{MyDarkRed}{rgb}{0.8,0.02,0.02}
\definecolor{MyDarkOrange}{rgb}{0.40,0.2,0.02}
\definecolor{MyPurple}{RGB}{111,0,255}
\definecolor{MyRed}{rgb}{1.0,0.0,0.0}
\definecolor{MyGold}{rgb}{0.75,0.6,0.12}
\definecolor{MyDarkgray}{rgb}{0.66, 0.66, 0.66}

%% file: figures/teaser/Teaser.tex
\setlength{\tabcolsep}{0pt}
\renewcommand{\arraystretch}{0}%
\begin{tabular}{@{}cc@{\hskip 2pt}c@{\hskip 2pt}ccccc@{}}%

  %\toprule%
  % Sample 1
  \multirow{2}{*}[30pt]{\includegraphics[height=60pt,width=60pt]{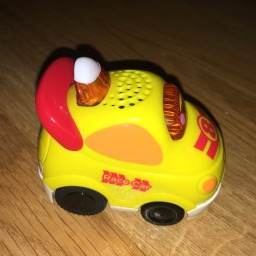}}%
                                                       & \includegraphics[height=30pt,width=30pt]{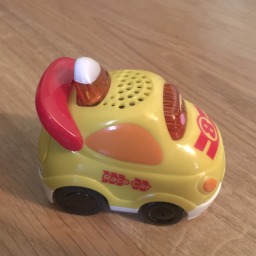}                                                                                                     %
                                                       & \multirow{2}{*}[30pt]{\includemedia[                                                                                                                                              %
      width=60pt, %
      height=60pt, %
      playbutton=none, %
      activate=pagevisible,%
      deactivate=pageinvisible,%
      addresource=figures/teaser/Toys_177_moving.mp4, %
      passcontext, %
  flashvars={source=figures/teaser/Toys_177_moving.mp4 & loop=true                                                                                                                          & autoPlay=true}                               %
    ]{\includegraphics[width=60pt,width=60pt]{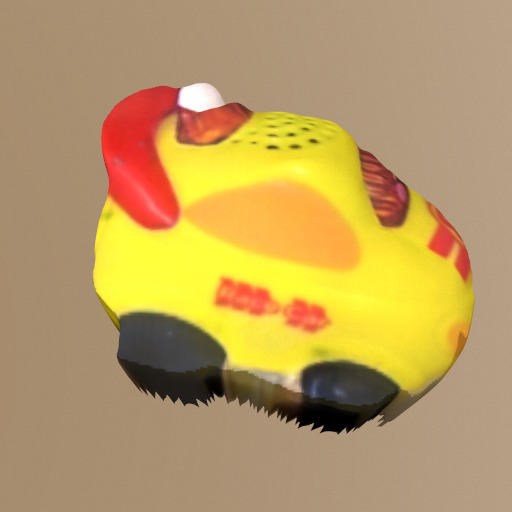}}{VPlayer.swf}} %
                                                       & \multirow{2}{*}[30pt]{\includegraphics[height=60pt,width=60pt]{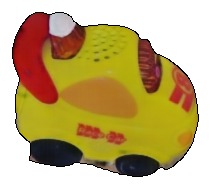}}                                                                      %
                                                       & \multirow{2}{*}[30pt]{\includegraphics[height=60pt,width=60pt]{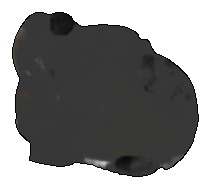}}                                                                     %
                                                       & \multirow{2}{*}[30pt]{\includegraphics[height=60pt,width=60pt]{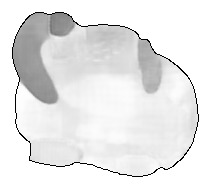}}                                                                    %
                                                       & \multirow{2}{*}[30pt]{\includegraphics[height=60pt,width=60pt]{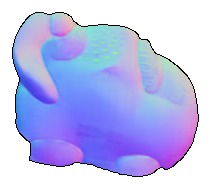}}                                                                       %
                                                       & \multirow{2}{*}[30pt]{\includegraphics[height=60pt,width=60pt]{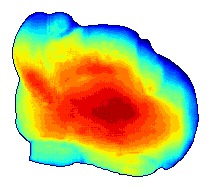}} \tabularnewline
                                                       & \includegraphics[height=30pt,width=30pt]{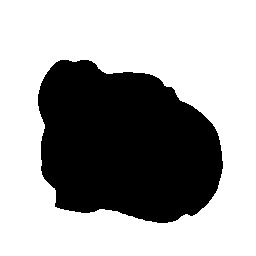}                                                     &                &  &  &  &  & \tabularnewline
  % Sample 2
  \multirow{2}{*}[30pt]{\includegraphics[height=60pt,width=60pt]{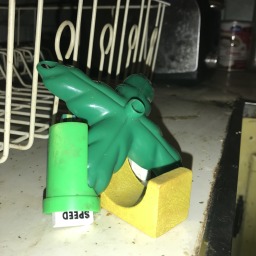}}%
                                                       & \includegraphics[height=30pt,width=30pt]{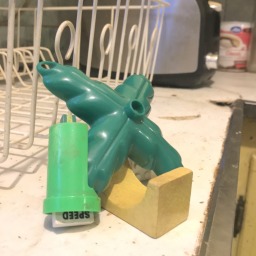}                                                                                                     %
                                                       & \multirow{2}{*}[30pt]{\includemedia[                                                                                                                                              %
      width=60pt, %
      height=60pt, %
      playbutton=none, %
      activate=pagevisible,%
      deactivate=pageinvisible,%
      addresource=figures/teaser/Toys_051_anim.mp4, %
      passcontext, %
  flashvars={source=figures/teaser/Toys_051_anim.mp4   & loop=true                                                                                                                          & autoPlay=true}                               %
    ]{\includegraphics[width=60pt,width=60pt]{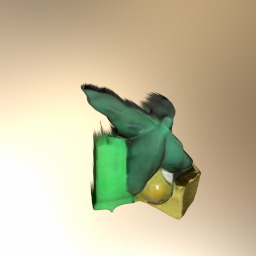}}{VPlayer.swf}} %
                                                       & \multirow{2}{*}[30pt]{\includegraphics[height=60pt,width=60pt]{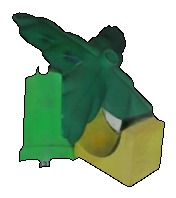}}                                                                      %
                                                       & \multirow{2}{*}[30pt]{\includegraphics[height=60pt,width=60pt]{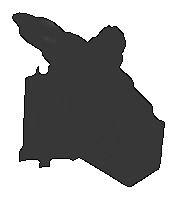}}                                                                     %
                                                       & \multirow{2}{*}[30pt]{\includegraphics[height=60pt,width=60pt]{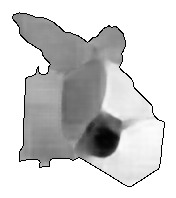}}                                                                    %
                                                       & \multirow{2}{*}[30pt]{\includegraphics[height=60pt,width=60pt]{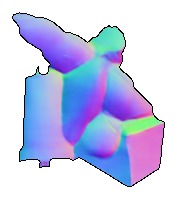}}                                                                       %
                                                       & \multirow{2}{*}[30pt]{\includegraphics[height=60pt,width=60pt]{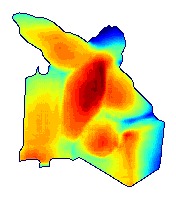}} \tabularnewline
                                                       & \includegraphics[height=30pt,width=30pt]{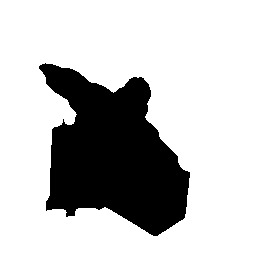}                                                     &                &  &  &  &  & \tabularnewline
  % Sample 3
  \multirow{2}{*}[30pt]{\includegraphics[height=60pt,width=60pt]{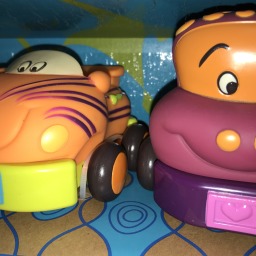}}%
                                                       & \includegraphics[height=30pt,width=30pt]{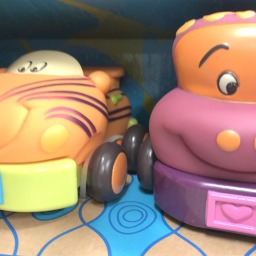}                                                                                                     %
                                                       & \multirow{2}{*}[30pt]{\includemedia[                                                                                                                                              %
      width=60pt, %
      height=60pt, %
      playbutton=none, %
      activate=pagevisible,%
      deactivate=pageinvisible,%
      addresource=figures/teaser/Toys_030_moving.mp4, %
      passcontext, %
  flashvars={source=figures/teaser/Toys_030_moving.mp4 & loop=true                                                                                                                          & autoPlay=true}                               %
    ]{\includegraphics[width=60pt,width=60pt]{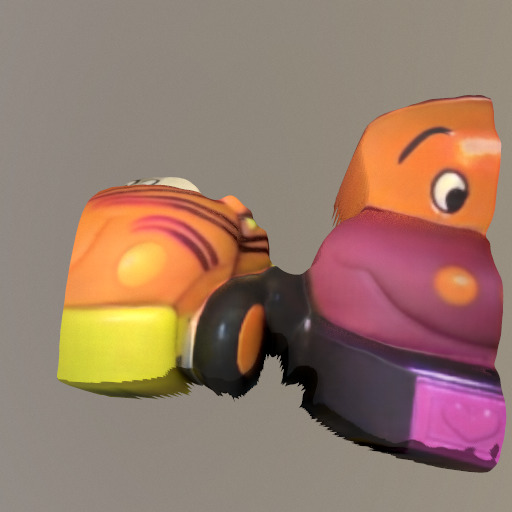}}{VPlayer.swf}} %
                                                       & \multirow{2}{*}[30pt]{\includegraphics[height=60pt,width=60pt]{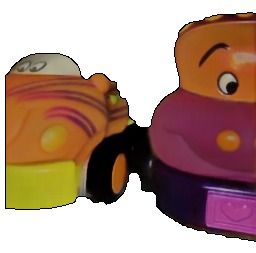}}                                                                      %
                                                       & \multirow{2}{*}[30pt]{\includegraphics[height=60pt,width=60pt]{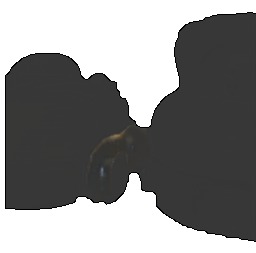}}                                                                     %
                                                       & \multirow{2}{*}[30pt]{\includegraphics[height=60pt,width=60pt]{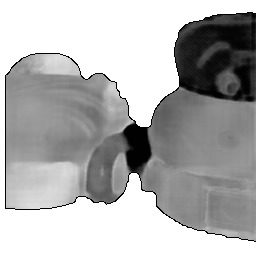}}                                                                    %
                                                       & \multirow{2}{*}[30pt]{\includegraphics[height=60pt,width=60pt]{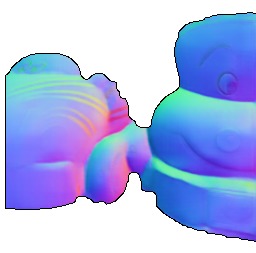}}                                                                       %
                                                       & \multirow{2}{*}[30pt]{\includegraphics[height=60pt,width=60pt]{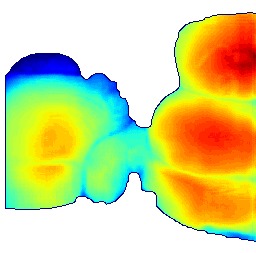}} \tabularnewline
                                                       & \includegraphics[height=30pt,width=30pt]{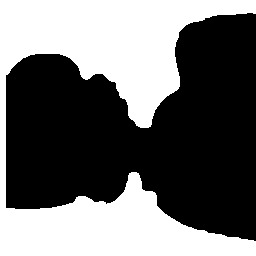}                                                     &                &  &  &  &  & \tabularnewline
  \midrule%
  \scriptsize Flash                                    & \pbox{30pt}{\scriptsize No-Flash                                                                                                                                                  \\Mask} & \scriptsize Novel Re-render & \scriptsize Diffuse & \scriptsize Specular & \scriptsize Roughness & \scriptsize Normal & \scriptsize Depth \\%
  \bottomrule%
\end{tabular}

%% file: sec_abstract.tex
%%%%%%%%% ABSTRACT
\begin{abstract}
\vspace{-4mm}
Capturing the shape and spatially-varying appearance (SVBRDF) of an object from images is a challenging task that has applications in both computer vision and graphics. %
Traditional optimization-based approaches often need a large number of images taken from multiple views in a controlled environment. %
Newer deep learning-based approaches require only a few input images, but the reconstruction quality is not on par with optimization techniques.  %
We propose a novel deep learning architecture with a stage-wise estimation of shape and SVBRDF. %
The earlier predictions guide each estimation, and a joint refinement network later refines both SVBRDF and shape. %
We follow a practical mobile image capture setting and use unaligned two-shot flash and no-flash images as input. %
Both our two-shot image capture and network inference can run on mobile hardware. %
We also create a large-scale synthetic training dataset with domain-randomized geometry and realistic materials. %
Extensive experiments on both synthetic and real-world datasets show that our networks trained on a synthetic dataset can generalize well to real-world images. %
Comparisons with recent approaches demonstrate the superior performance of the proposed approach. %

% We propose a multi-stage deep learning approach that bridges the gap between traditional optimization and deep learning approaches by first estimating geometry followed by illumination and appearance (SVBRDF) estimation. The previous predictions guide each estimation. Lastly a joint refinement network 
% % is performed which optimizes 
% refines the predicted geometry and SVBRDF at the same time. We use unaligned two-shot flash and no-flash images as input and introduce a new large scale synthetic dataset for training. Due to our architectures uses smaller networks. All models can be converted to run on mobile devices. This enables on-device capture, prediction, and re-rendering of objects. Experiments on synthetic and real-world datasets show improvements over state-of-the-art SVBRDF and shape estimation methods.

\end{abstract}

%% file: sec_introduction.tex
\vspace{-4mm}
\section{Introduction}
\label{sec:intro}
\vspace{-2mm}

The estimation of intrinsic attributes of a scene such as shape and reflectance of objects and the illumination condition of the scene is often called as an \emph{inverse rendering problem} in computer vision~\cite{Sengupta2019, Ramamoorthi2001, Kimiccv17}, and has been a core of many applications such as relighting of images~\cite{Ren15}, photo-realistic mixed reality~\cite{Meka2018}, and automatic creation of assets for content creation tasks~\cite{Aittala2015}. %

In this work, we are interested in the automatic estimation of the shape and appearance of the object in a scene from only two images. %
In particular, we represent the shape of the object with a depth map and the appearance as a Bidirectional Reflectance Distribution Function (BRDF)~\cite{Nicodemus1965}. %
A BRDF describes the low-level material properties of an object that defines how light is reflected at any given point on an object surface. %
One of the most popular parametric models~\cite{Cook1982} represents the diffuse and specular properties and the roughness of the surfaces. %
Since the material properties can vary across the surface, one has to estimate the BRDF at each image pixel for a more realistic appearance (i.e., spatially-varying BRDF (SVBRDF)). %

As the BRDF is dependent on view and light directions and estimating depth from a single 2D image is an ambiguous task, multi-view setups improve the estimation accuracy of both shape~\cite{schonberger2016structure} and BRDF~\cite{Meka2018}. %
Predicting shape and BRDF from only a few images is still very challenging. %
For shape estimation, the advances in deep learning-based depth estimation allow us to estimate the depth of a single~\cite{fu2018deep, Lasinger2019}, or a pair of images~\cite{Ummenhofer17cvpr} efficiently. %
As monocular depth estimation is not as accurate as multi-view approaches, we exploit shading cues on the surface to disambiguate the geometric shape~\cite{alldrin:2008a, Zhang-sfs-99} in our approach. %

%%% Aim of this work.
We propose a neural network-based approach to estimate SVBRDF and shape of an object along with the illumination from given two-shot images: flash and no-flash pairs. %
Some recent deep learning approaches~\cite{Deschaintre2018, Li2018, Li2018a} for BRDF estimation use only a single 
% LDR 
flash image as input. %
Flash images often have harsh reflective highlights where the input pixel information is saturated in non-HDR images. %

Li \etal~\cite{Li2018a} uses a single input image and estimates shape and part of the BRDF, such as diffuse albedo and the roughness while ignoring the specular color. %
In this work, we use flash and no-flash image pairs as input allowing the network to access pixel information from the no-flash image when the corresponding pixels are saturated in the flash image. %
We focus on practical utility: Our input capture setup follows a real-world scenario where the two-shot images are consecutively taken using a mobile phone camera in burst capture. %
The system is designed to tackle the 
% to be robust against 
misalignment between the two-shot images due to camera shake.% between the two images. %

%%% Challenge1: Availability of training data. How we tackle this with synthetic data?
A pivotal challenge for any learning approach is the need for training data. %
We tackle this issue by creating a large-scale synthetic dataset. %
Flash and no-flash images are rendered using high-quality, human-authored SVBRDF textures that are applied to synthetic geometry generated by domain randomization~\cite{Tobin2017} of geometric shapes and backgrounds. %
Our networks trained on this synthetic data generalize well to real-world object images. %

%%% Challenge2: Problem ambiguity. How we tackle this with a cascaded network?
Another key challenge in shape and SVBRDF estimation is the problem of ambiguity. %
For example, a darker region in an image could be created by its material color being dark, the area slightly shadowed due to its shape, or the illumination at that spot being darker. %
We tackle this ambiguity by using a cascaded approach, where separate neural networks are used to estimate shape (depth), illumination, and SVBRDF. %
Specifically, we first estimate depth and normals using a geometry estimation network. %
Then the illumination is approximated, followed by SVBRDF reconstruction. %
Each step is guided by the estimates of the previous networks. %
Finally, shape and SVBRDF are optimized jointly using a refinement network. %
Each task is implemented by specialized network architectures. %
Empirically, this cascaded regression approach works reliably better compared to a single-step joint estimation. %
As a favorable side-effect of this cascaded approach, the size of each network is small compared to a large joint estimation network. %
This allows the inference networks to even operate on a mobile device. %
Coupled with two-shot mobile capturing, this presents a highly practical application. %

%%% Briefly about experiments and take-home messages.
Quantitative analysis based on a synthetic dataset comprising of realistic object shapes and SVBRDFs demonstrates that our approach produces more accurate estimates of shape and SVBRDF compared to baseline approaches. %
We also qualitatively demonstrate the applicability of our approach on a real-world two-shot dataset~\cite{Yagiz2018}.

%% file: sec_related.tex
\begin{figure*}[t!]
    \centering
    \includegraphics[width=0.9\linewidth]{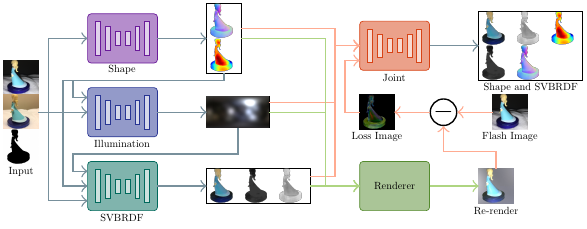}
    \vskip-3mm
    \mycaption{Cascaded Network}{Overview of the inference pipeline for shape, illumination and SVBRDF estimation.}
    \label{fig:pipeline_overview}
\end{figure*}

\section{Related work}
\vspace{-2mm}

The literature on object SVBRDF and/or shape estimation is vast. Here, we only discuss the representative works that are related to ours.

\vspace{1mm}
\noindent \textbf{BRDF Estimation} %
An exhaustive sampling of each BRDF dimension demands long acquisition times. %
Several proposed methods focus on reducing acquisition time~\cite{Lensch2003, Aittala2013, Dong2010}. %
These methods introduce capture setups and optimization techniques that reduce the number of images required to reconstruct high-quality SVBRDF. %
Recently, several attempts~\cite{Deschaintre2018, Li2017, Li2018, Aittala2018, Aittala2015} reconstruct the SVBRDF on flat surfaces with one or two flash images. %
These approaches leverage neural networks trained on large amounts of data and resolve the problem of ambiguity to some extent by learning the statistical properties of BRDF parameters. %

For a joint estimation of shape and shading, separate optimization steps for shape and shading are common~\cite{Lensch2001,Nam2018, Goldman2009, Barron2015}. % 
Lensch \etal~\cite{Lensch2001} introduce Lumitexels, which stack previously acquired shape information with the luminance information from the input images, to guide the BRDF estimation and to reduce ambiguities in the optimization. %
Compared to a joint estimation, fewer local minima are found, and the optimization is more robust. %
Recently, the task of predicting the shape and BRDF of objects or scenes is also addressed using deep learning models~\cite{Li2018a, Sengupta2019}. %
Li \etal~\cite{Li2018a} predict the shape and BRDF of objects from a single flash image using an initial estimation network followed by several cascaded refinement networks. %
Here, the BRDF consists of diffuse albedo and specular roughness but lacks the specular albedo. %
Specularity is, however, essential in re-rendering metallic objects, for example. %

Compared to Li \etal~\cite{Li2018a}, our method additionally estimates the SVBRDF with specular albedo. %
In comparison to flat surface SVBRDF estimation~\cite{Deschaintre2018, Li2018, Aittala2018, Aittala2015}, our method handles full objects with shape from any view position. %
Additionally, due to our unaligned two-shot setup, saturated flash highlights are better compensated, while still providing the same one-button press capture experience for the user, due to our mobile capture scenario. %

\vspace{1mm}
\noindent \textbf{Intrinsic Imaging} %
Intrinsic imaging is the task of decomposing an image of a scene into reflectance (diffuse albedo), and shading~\cite{Barrow1978,Barron2015, maier2017, Tappen2005}. %
With the advance in deep learning, the problem of separating shape, reflectance, and shading is tackled from labeled data~\cite{lettry2018, Narihira2015, Shi2017}, unlabeled~\cite{LiIntrinsic2018} and partially labeled data~\cite{Zhou2015, LiIntrinsicCg2018, Nestmeyer2017, Bell2014}. %
Due to the very simplistic rendering model, the use cases are limited compared to our SVBRDF estimation setup, which can be used for general re-rendering in new light scenarios. %

\vspace{1mm}
\noindent \textbf{Shape Estimation} %
One can obtain high-quality depth from stereo images, but the problem of monocular depth estimation is quite challenging.
% The task of shape estimation from rectified stereo images is mostly solved. % 
%For monocular depth estimation, this problem is recently tackled with the rise of deep learning
Monocular depth estimation is predominantly tackled with deep learning~\cite{wang2015, liu2016, Li2019, Huan2018, Roy2016, Lasinger2019} in the recent years. %
This problem is especially challenging as no absolute scale is known from single images, and the depth cues need to be resolved by shading information such as the quadratic light fall-off~\cite{liao2007}.

%% file: sec_approach.tex
% \begin{figure*}[t!]
%     \centering
%     \includegraphics[width=0.9\linewidth]{figures/OverviewTikz.pdf}
%     \vskip-3mm
%     \mycaption{Cascaded Network}{Overview of the Inference Pipeline of the cascaded Network. \hlnote{move to top of page 3 instead of 4}}
%     \label{fig:pipeline_overview}
% \end{figure*}

\section{Methods}
\label{sec:method}
\vspace{-2mm}

As briefly discussed in the introduction, to tackle the problem of ambiguity in shape and SVBRDF estimation, we propose a novel cascaded network design for shape, illumination, and SVBRDF predictions. %
\fig{fig:pipeline_overview} shows an overview of our cascaded network. %

\noindent \textbf{Problem Setup}
Our network takes two-shot object images (flash and no-flash) with the corresponding foreground object mask and estimates shape and SVBRDF. %
We also estimate illumination as a side-prediction to help shape and SVBRDF prediction. %
The two-shot images can be slightly misaligned to support practical image capture with a handheld camera. %
The object mask allows us to evaluate only the pixels of the object in the flash image and is easily generated with GrabCut~\cite{rother2004}. %
The object shape is represented as depth and normal at each pixel. %
The depth map provides a rough shape of the object, while the normal map models local changes more precisely. %
This shape representation is commonly used in various BRDF estimation methods~\cite{Li2018a, Nam2018}. %
We use the Cook-Torrence model~\cite{Cook1982} to represent the BRDF at each pixel with diffuse albedo (3 parameters), specular albedo (3), and roughness (1). %
Similar to \cite{wang2018, li2019sg}, we estimate the environment illumination with 24 spherical Gaussians. %
%In summary, the network takes two-shot images and object masks as input and outputs depth, normals, diffuse albedo, specular albedo, and roughness at each pixel. %
%As a side output, we also predict spherical Gaussian illumination. %

\vspace{1mm}
\noindent \textbf{Network Overview and Motivation} %
In order to tackle the shape/SVBRDF ambiguity, we take the inspiration from traditional optimization techniques~\cite{Lensch2001,Nam2018}, which iteratively minimize a residual and alternate between optimizing for shape and/or reflectance. %
Thus, separate networks are used for shape, illumination, and SVBRDF estimation in a cascaded as well as an iterative manner. %
Predictions from earlier stages of the networks in the cascade are used as inputs to later networks to guide network predictions to better solutions. %
In addition, the scene is re-rendered with the current estimates, and refined further using the residual image. %

Since flash and no-flash images are slightly misaligned, shape estimation is less challenging compared to SVBRDF estimation. %
Mis-alignment in pixels, as well as pixel differences between two-shot images~\cite{liao2007}, are a good indicator of object depth. %
Thus, we first predict depth and normals using a specialized merge convolutional network followed by a shape-guided illumination estimation. %
Then, the SVBRDF is predicted with the current estimates of shape and illumination as additional input. %
Finally, after computing a residual image, we refine both shape and SVBRDF using a joint refinement network. Refer to the supplementary for network architecture details.

\subsection{Shape Estimation with Merge Convolutions}
\label{sec:shape_estimation}
\vspace{-1mm}
Since the camera parameters are unknown and the two-shot images have a minimal baseline, traditional struc\-ture-from-motion or stereo solutions are not useful for dense depth estimation. %
The shape estimation needs to rely on the unstructured perspective shift as well as pixel differences between flash and no-flash images. %
In order to tightly integrate information from both the images, we design a specialized convolutional network for shape estimation. %

\begin{figure}
    \centering
    \includegraphics[width=0.98\linewidth]{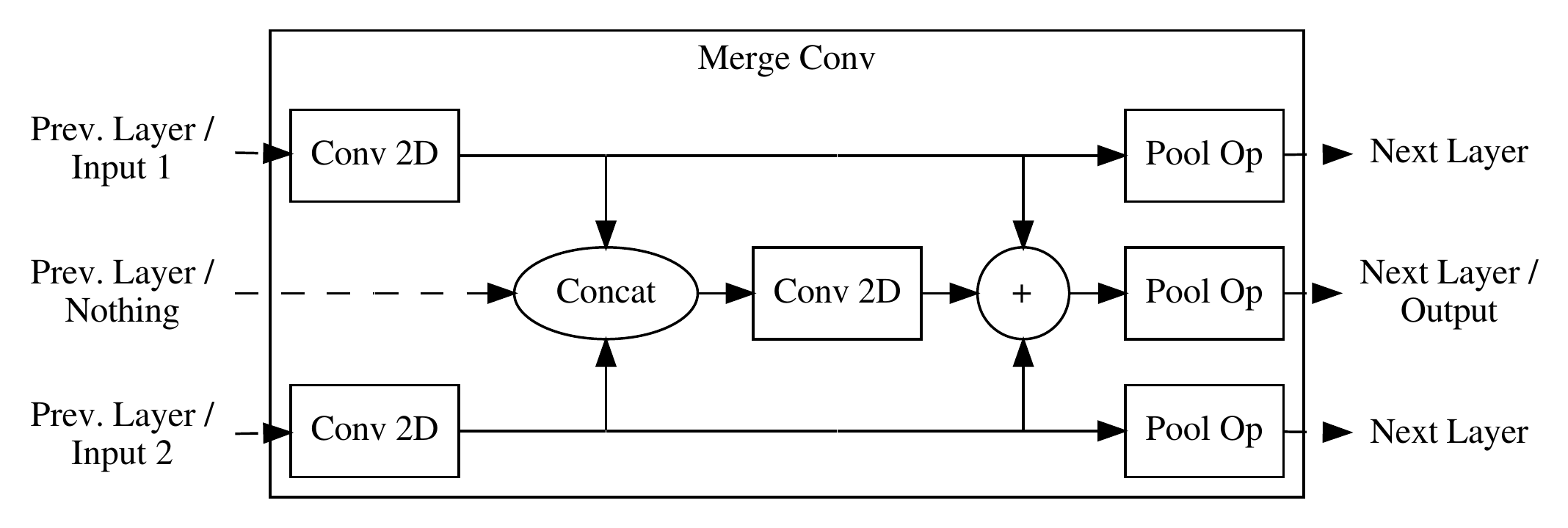}
    \vskip-2mm
    \mycaption{Merge convolutions}{The merge convolution provides separate pathways for the two-shot inputs and merges the information in a third path.}
    \label{fig:merge_conv}
\end{figure}

For depth and normal map prediction, we use a U-net like encoder-decoder architecture~\cite{Ronneberger2015}. %
Instead of standard convolution blocks, we propose to use novel merge convolution blocks (MergeConv). %
We concatenate the object mask with each of the two-shot input images as input to the network. %
\fig{fig:merge_conv} illustrates the MergeConv block. %
Both the input images or their intermediate features are separately processed by 2D convolutions (Conv2D). %
The outputs of each Conv2D operation are concatenated in channels with the merged output from the previous MergeConv layer and is processed with another Conv2D operation. %
Inspired by residual connections in ResNet~\cite{He2016}, we add the Conv2D outputs as indicated in \fig{fig:merge_conv}. %
We use 4 MergeConv blocks for the encoder and also 4 for the decoder. %
During encoding, max pooling for 2$\times$ spatial downsampling is used. %
% Furthermore, 
For each MergeConv in the decoder, we use 2$\times$ nearest neighbor upsampling. %
The final depth and normal map estimates are produced using a separate 2D convolution, followed by a sigmoid activation. %
The rationale behind this MergeConv architecture is to keep separating the process of pathways for both the input images while exchanging (merging) the information between them using a third pathway in the middle. %
We believe that information in both input images is essential for shape reasoning, and this architecture helps to keep the features from each of the images intact throughout the network. %
Empirically, we observe reliably better shape predictions with this architecture compared to a standard U-net with a similar number of network parameters. %

Training losses are based on the $\mathcal{L}_2$ distance between ground-truth (GT) and predicted depths, $\mathcal{L}^\text{depth}_2$, as well as the angular distance between GT and predicted normals, $\mathcal{L}^\text{normals}_\textit{angular}$. %
Besides, we use a new consistency loss between the predicted normal $\mathbf{n}$ and a normal $\mathbf{n}^*$ derived from the depth  information $\mathbf{d}$, which enforces that the predicted normals follow the curvature of the shape:
\begin{align}
\vspace{-3mm}
    \mathcal{L}^{\text{normals/depth}}_\textit{consistency} &= \frac{\mathbf{n}}{\left\lVert \mathbf{n} \right\lVert} - \frac{\mathbf{n}^*}{\left\lVert \mathbf{n}^* \right\lVert}, \label{eq:consistency} \\
    \mathbf{n}^* = \begin{bmatrix}
        \triangledown \mathbf{d}  & 2\frac{1}{\text{width}}
    \end{bmatrix}^T &= \begin{bmatrix}
        \frac{\partial \mathbf{d}}{\partial x} & \frac{\partial \mathbf{d}}{\partial y} & 2\frac{1}{\text{width}}
    \end{bmatrix}^T, \label{eq:height_derived_normals}
\end{align}
The normal $\mathbf{n}^*$ is derived from the depth map using gradients along horizontal ($x$) and vertical ($y$) directions. %
The $z$ component can be considered a strength factor which is derived from the image width. %
The total loss is a weighted combination of the three losses: 
$\mathcal{L}^\text{depth}_2 + \mathcal{L}^\text{normals}_\textit{angular} + 0.5 \times  \mathcal{L}^\text{normals/depth}_\textit{consistency}$.

\subsection{Shape Guided Illumination Estimation} %
\vspace{-1mm}
To guide SVBRDF predictions, we also estimate the environment illumination. %
Hereby, the BRDF prediction can consider environment light and reduce additional highlights as well as improve the albedo colors and intensities.
The illumination is represented with 24 spherical Gaussians (SG), where each SG is defined by amplitude, axis, and sharpness. %
However, we only estimate the amplitude and set the axis and sharpness to cover a unit sphere. %
The estimation thus only estimates the amplitudes of the SG resulting in 24 RGB values. %
As the environment illumination can reach very high values and the flash and no-flash input images are in LDR, 
SG amplitudes are constrained to values between 0 and 2.
% a value range of 0-2 for the SG amplitudes is allowed. %
Refer to the supplementary for environment map samples and their SG representations. %

We use a small convolutional encoder network followed by fully-connected layers for illumination estimation. The network receives two-shot images, object mask, and the previously predicted depth and normals as input. %
As illumination is reflected on the surface towards the viewer, the previously estimated shape information helps in better illumination estimations. %
To train the illumination network, we use the $\mathcal{L}_2$ distance between predicted and ground-truth SGs as the loss function. %

\subsection{Guided SVBRDF Estimation} %
\label{subsec:svbrdf_network}
\vspace{-1mm}
SVBRDF estimation becomes a less ambiguous task when conditioned on known object shape and environment illumination. %
Thus, together with two-shot images, the previously estimated depth, normals, and illumination are used as input to the SVBRDF network to predict diffuse albedo and specular color as well as surface roughness at each pixel. %
Following recent work on BRDF estimation~\cite{Li2017,Li2018, Deschaintre2018}, the U-net architecture~\cite{Ronneberger2015} is used in our SVBRDF network. %

\vspace{1mm}
\noindent \textbf{Differentiable Rendering} %
We develop a differentiable rendering module to re-render the object flash image from the estimated depth, normals, illumination, and SVBRDF. %
At each surface point, the renderer evaluates the direct light from the flash-light source and the estimated environment illumination and integrates it with the BRDF to compute the reflected light~\cite{Kajiya1986}.  %
Fast evaluation of the environment illumination is achieved by representing the illumination as well as the BRDF model as spherical Gaussians (SG)~\cite{Wang2009}. %
The product of two SGs is an SG, and the integral of an SG has a closed-form solution that is inexpensive to compute.
% an inexpensive to compute closed-form solution. %

\begin{figure*}
    \centering
    \begin{subfigure}[b]{0.15\textwidth}
         \centering
         \includegraphics[width=\textwidth]{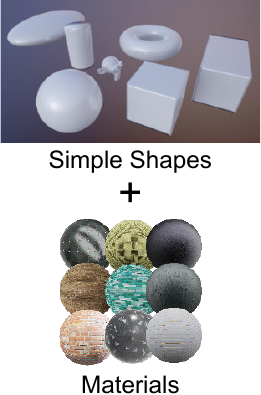}
     \end{subfigure}
     \hfill
    \begin{subfigure}[b]{0.84\textwidth}
         \centering
         \includegraphics[width=\textwidth]{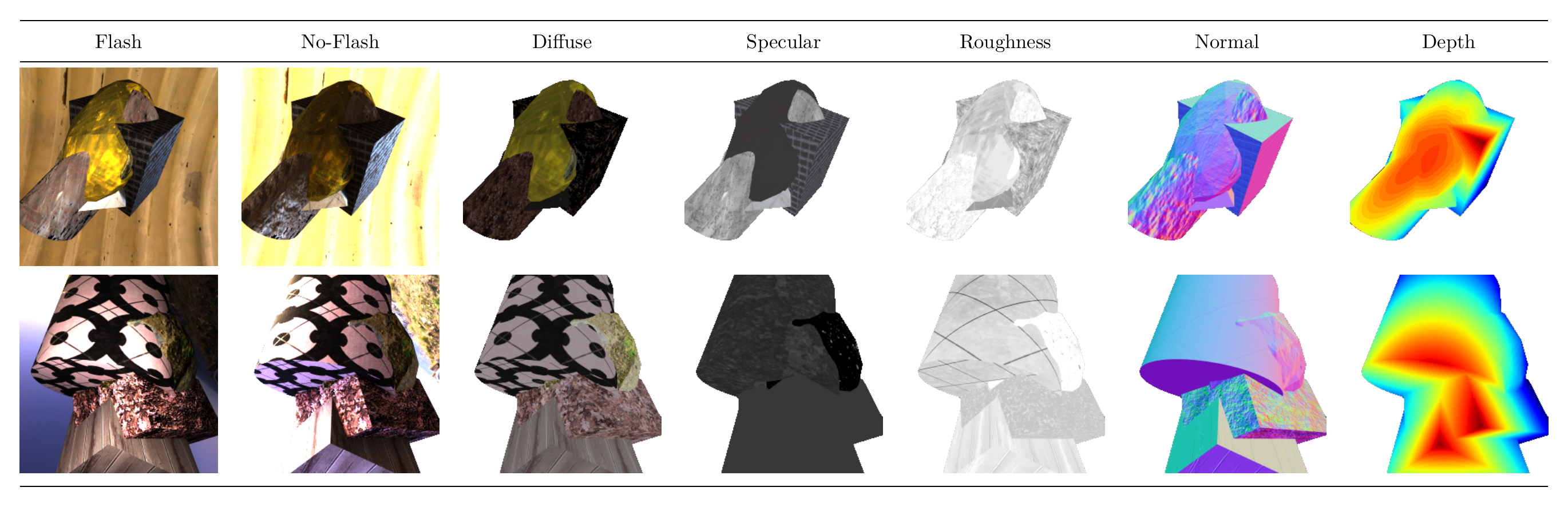}
     \end{subfigure}
     \mycaption{Large-scale Synthetic Dataset}{ (Left) Samples of primitive shapes and materials used for the dataset creation, (Right) The visualization of two examples with various properties.}
     \label{fig:large_scale_dataset}
\end{figure*}

\vspace{1mm}
\noindent \textbf{Loss Functions for SVBRDF Network} %
The SVBRDF network is trained using a combination of different loss terms: the mean absolute error (MAE) between GT and the predicted SVBRDF parameters as well as a loss between a synthetic direct illumination only flash GT image and re-rendered direct illumination flash image. %
The rendering loss is back-propagated through the differentiable renderer to update the SVBRDF network. %
As rendering can result in large values from specular highlights, the MAE loss is calculated on $\log(1+x)$, where $x$ refers to the direct light only synthetic input and the re-rendered image. %

\subsection{Joint Shape and SVBRDF Refinement} %
\vspace{-2mm} %
In our cascaded network, we use the estimated depth to guide the SVBRDF prediction. %
Likewise, one can obtain better depth prediction with known SVBRDF. %
We jointly optimize depth, normals, and SVBRDF using a separate refinement network. %
For this refinement, all the earlier predictions along with the residual loss image between the re-rendered previous result and the input flash image are used. %
%Given the residual image, and the previous predictions, the refinement network predicts refined depth, normals, and SVBRDF parameters. %
The network architecture is a small CNN encoder and decoder of 3 steps, each with 4 ResNet blocks \cite{He2016} in-between. %
The loss function is an MAE loss between the predicted parameter maps and ground truth ones. %

\subsection{Implementation} %
\vspace{-2mm}
The cascaded networks along with the differentiable renderer are implemented in Tensorflow~\cite{tensorflow2015}. %
The overall pipeline consists of 4 networks, as illustrated in \fig{fig:pipeline_overview}. %

\vspace{1mm}
\noindent \textbf{Runtime} %
%Each of the networks is relatively small, and the overall pipeline takes 700 ms, including rendering for inference on a 256$\times$256 image on an Nvidia 1080 TI GPU. On a Google Pixel 4 mobile device, the evaluation takes roughly 6 seconds. %
%The rendering step takes about 220ms on a single-threaded desktop CPU (AMD Ryzen 7 1700) and similar speeds on Google Pixel 4. %
%Refer to the supplementary for further runtime analysis. %
Each of the networks is relatively small, and the overall inference pipeline takes 700ms on a 256×256 image on an NVIDIA 1080 TI, including the required rendering step. %
On a Google Pixel 4, the evaluation takes roughly 6 seconds. %
The rendering step is implemented in software and takes 220ms on a single-threaded desktop CPU (AMD Ryzen 7 1700) and similar speeds on a Google Pixel 4. %
%Refer to the supplementary for further analysis. 

\vspace{1mm}
\noindent \textbf{Training} %
All the networks are trained for 200 epochs with 1500 steps per epoch using the ADAM optimizer~\cite{kingma2014adam} 
with a learning rate of $2\mathrm{e}{-4}$ at the beginning, which is reduced by half after 100 epochs. %
The networks are trained sequentially as each network in the cascade uses the result of earlier networks as input. %

\vspace{1mm}
\noindent \textbf{Mobile Application for Scene Capture and Inference} %
In addition to producing better results, another major advantage of the cascaded network design compared to a single joint network is that each of the sub-networks is small, and the overall network can fit on mobile hardware. %
We convert the network models to Tensorflow Lite that runs on mobile hardware and develop a highly practical android application that can successively capture two-shot flash and no-flash images and runs the cascaded network to estimate SVBRDF and shape. %
We use on-device GrabCut~\cite{rother2004} to obtain the object mask. %
In \fig{fig:real_world_mobile} a prediction from the mobile application is shown. %
Refer to supplementary for more details on the mobile application and further predictions. %

\section{Large-scale SVBRDF \& Shape Dataset}
\label{sec:dataset}
\vspace{-2mm}
It is very time consuming and expensive to scan SVBRF of real-world objects. %
Since we rely on deep learning techniques for SVBRDF and shape estimation, vast amounts of data are needed for network supervision. %
We create a large-scale synthetic dataset with realistic SVBRDF materials. %

\vspace{1mm}
\noindent \textbf{High-quality Material Collection} %
We gather a collection of publicly available human-authored, high-quality SVBRDF maps from various online sources~\cite{3dtextures, cc0textures, cgbookcase, freepbr, sharetextures,texturehaven}. %
The parameterization of these collected SVBRDF maps is for the Cook-Torrence model~\cite{Cook1982}. %
In total, the collection consists of 1125 high-resolution SVBRDF maps. %
To further increase the material pool, we randomly resize and take $768 \times 768$ crops of these material maps. %
We additionally apply random overlays together with simple contrast, hue, and brightness changes. %
%In the end, 11,250 material maps are available in the dataset. %
The final material pool contains 11,250 material maps. %
Sample material maps are shown in \fig{fig:large_scale_dataset}. %

\vspace{1mm}
\noindent \textbf{Domain Randomized Object Shapes} %
One option for generating 3D objects is to gather realistic object meshes and apply materials to those. %
However, it is challenging to collect large-scale object mesh data covering a wide range of object categories. %
Moreover, mapping the object meshes to the corresponding materials (\eg, using ceramic materials for teapots) would result in a small dataset, and thus, applying random materials to object meshes is a reasonable strategy. %
We notice that applying random material maps to complex-shaped object meshes would result in distorted texture or tiling artifacts. %
Because of these numerous challenges, we choose to randomize object shapes to synthesize large-scale data. %
Following Xu \etal~\cite{xu2018}, a randomly chosen material is applied to 9 different shape primitives such as spheres, cones, cylinders, tori, etc. %
We randomly choose 6 to 7 material-mapped primitive shapes and place them randomly to assemble a scene. Sample object shape primitives are shown in \fig{fig:large_scale_dataset}. %
This strategy is similar to domain randomization~\cite{Tobin2017} (DR) that is shown to be useful in high-level semantic tasks such as object detection~\cite{tremblay2018training}. %
Here, we demonstrate the use of DR for the low-level yet complex task of SVBRDF and shape estimation. %
For simplicity, we refer to our material-mapped and geometry randomized object shapes as DR objects. %
\fig{fig:large_scale_dataset} shows sample primitive shapes, materials and resulting DR objects with GT shape and SVBRDF parameters. %

\vspace{1mm}
\noindent \textbf{HDR Illumination} %
For environment illumination, we collect 285 high-dynamic-range (HDR) illumination maps from~\cite{hdrihaven}. %
These maps are images in latitude-longitude format, which are wrapped on the inside of a sphere, which acts as a light source for the DR object. %

\vspace{1mm}
\noindent \textbf{Rendering} %
We use the Mitsuba~\cite{mitsuba} renderer to create two-shot flash and no-flash images of a DR object illuminated with a randomly chosen illumination. %
In total,  the DR dataset contains 100K generated scenes. %
Note that each DR object consists of differently sampled primitive shapes, and the distance of the closest surface from the camera varies across different DR objects. %
This setup mimics the real-world capture setting where the object distance to the camera varies. %
For the no-flash image rendering, the camera position is slightly shifted to mimic the camera shake in a mobile scene capture. %

In addition to the two-shot flash and no-flash images, we also render another flash image that only has direct illumination. %
This direct illumination flash image is used to additionally supervise the SVBRDF network after differentiable rendering (\sect{subsec:svbrdf_network}). %
This direct illumination only image is solely used for training supervision and is not required for inference. Besides, we render GT depth, normals, diffuse albedo, specular albedo, and roughness maps, using Mitsuba~\cite{mitsuba}, that are used for direct network supervision. %
\fig{fig:large_scale_dataset} shows samples from this dataset with more in the supplementary, which also provides additional details on the rendering setup. %

%% file: sec_results.tex
\section{Experiments}
\label{sec:experiments}
\vspace{-2mm}

We evaluate our approach on both synthetic and real datasets and compared with several baseline techniques.
% Evaluating the performance of the cascaded network architecture is shown in several experiments. %
In this section, we present both quantitative and qualitative results and refer to the supplementary materials
for further visual results and comparisons. %

\begin{figure*}
    \centering
    \input{figures/evaluation/syn_li_comparison/comparison.tex}
    \mycaption{Comparison with Li \etal~\cite{Li2018a}}{Ours estimates the diffuse, depth and normal more accurately in particular.}
    \label{fig:li_comparison_example}
\end{figure*}

\begin{figure}
    \centering
    \includegraphics[width=0.85\linewidth]{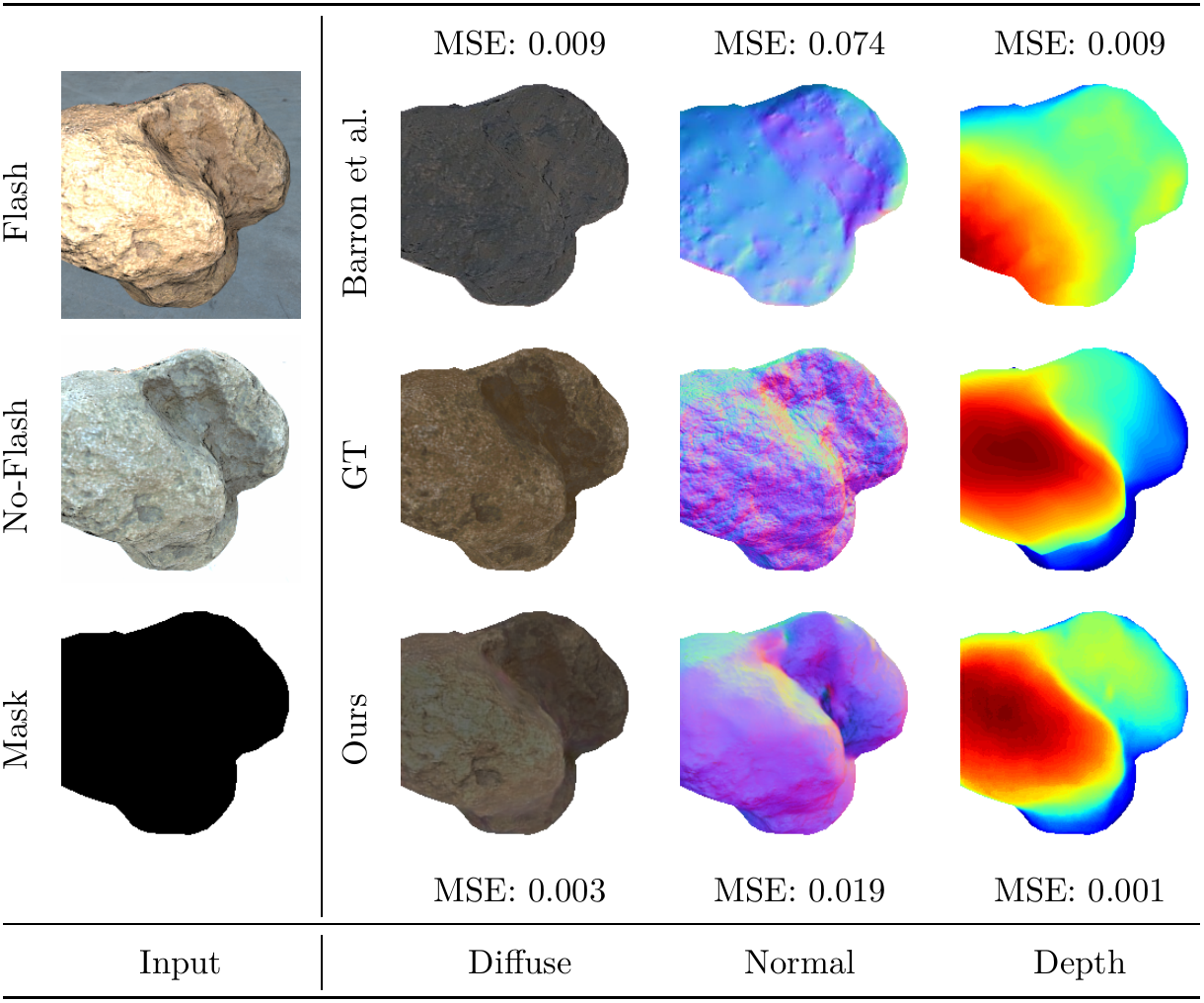}
    \mycaption{Comparison with Barron \etal~\cite{Barron2015} (SIRFS)}{Barron \etal does not estimate specular and roughness parameters.}
    \label{fig:barron_comparison}
\end{figure}

\begin{figure*}
    \centering
    \input{figures/evaluation/real_world_li_comparison/RealWorldParams.tex}
    \mycaption{Real-world comparison}{Comparison with Li \etal~\cite{Li2018a} on a real-world sample from \cite{Yagiz2018}.}
    \label{fig:real_world_comparison}
\end{figure*}

\begin{figure*}
    \centering
    \input{figures/evaluation/real_world_own/RealMobileExample.tex}
    \mycaption{Mobile capture and inference}{A result from our mobile  application that does two-shot image capture followed by SVBRDF and shape estimation.}
    \label{fig:real_world_mobile}
\end{figure*}

\vspace{1mm}
\noindent \textbf{Test datasets} %
We quantitatively validate the proposed method on synthetic data with realistic object shapes and SVBRDF and also qualitatively on a real-world two-shot image dataset~\cite{Yagiz2018}. %
Images of both of these datasets are unseen during network training. %
For synthetic test data, we collected 20 freely available, fully textured 3D objects with realistic shapes and materials~\cite{cgtrader}. %
These objects are rendered using the Mitsuba renderer~\cite{mitsuba} with unseen HDR illumination maps. %
\fig{fig:li_comparison_example} and~\ref{fig:barron_comparison} show samples of two-shot input images of our synthetic test dataset. %

For real-world evaluation, we use two-shot images from the recent 'flash and ambient illuminations dataset' from~\cite{Yagiz2018}. %
We have created foreground object masks on several samples from the 'Objects' and 'Toys' category, as these fit the single object assumption. %
This dataset does not contain ground truth BRDF parameters, but the visual quality can be inspected on the estimations and also on re-renderings with different camera views and illuminations. %

\vspace{1mm}
\noindent \textbf{Metrics} %
To evaluate the quality of the shape and SVBRDF predictions, we mainly use metrics that directly compare the ground truth (GT) and predictions. %
For the depth and normal estimations, a Mean Square Error (MSE) is a fitting candidate. %
To enable comparisons with methods that predict relative depths, we employ a Scale-Shift Invariant Metric as in~\cite{Lasinger2019}. %
Refer to the supplementary for details. %
For SVBRDFs, there exists no clear metric which aligns with human perception of materials. %
Following previous works, we also use the MSE metric on SVBRDF prediction maps. %

\begin{table}
    \centering
    \resizebox{0.95\linewidth}{!}{
    \begin{tabular}{@{} rcccccc @{}}
    \toprule
    Parameter & MiDaS~\cite{Lasinger2019} & SIRFS~\cite{Barron2015} & RAFII~\cite{Nestmeyer2017} & Li \etal~\cite{Li2018a} & Ours-JN & Ours-CN \\ \midrule
    Diffuse     & NA      & [0.033] & \textbf{[0.018]} & 0.160/[0.019] & 0.065/[0.022] & \textbf{0.060}/[\textbf{0.018}] \\
    Specular    & NA      & NA      & NA & NA            & 0.053         & \textbf{0.047} \\
    Roughness   & NA      & NA      & NA & 0.072         & 0.064         & \textbf{0.061} \\
    Normal      & NA      & 0.089   & NA & 0.034         & 0.025         & \textbf{0.021} \\
    Depth       & [0.006] & [0.021] & NA & [0.024]       & [0.005]       & [\textbf{0.004}] \\
    \bottomrule
    \end{tabular}}
    \\[2pt]
    \mycaption{State-of-the-art comparison}{The Mean Square Error (MSE) on a sample dataset of 20 unseen objects. Scale and shift invariant metric in [.] where it applies. For the diffuse color this metric is only scale invariant.}
    \label{tbl:comparison}
\end{table}

\subsection{Ablation Study}
\label{subsec:ablation}
\vspace{-2mm}
Within our framework, we empirically evaluate different choices we make in our network design. %

\vspace{1mm}
\noindent \textbf{Cascade vs. Joint Network} %
We compare our cascaded network with a single large joint network that estimates all the shape and SVBRDF parameters together. %
For a fair comparison, we design a joint  (JN) that has a comparable number of network parameters as our cascaded network (CN) (`Ours-CN' vs.\ `Ours-JN'). %
The JN follows the U-Net~\cite{Ronneberger2015} architecture. %
\tbl{tbl:comparison} shows the quantitative comparisons between them. %
Results indicate that the CN consistently outperforms JN on both SVBRDF and shape estimations, by a significant margin. %
This empirically underlines the usefulness of our guided stage-wise estimation and joint refinement compared to using a single large network for joint SVBRDF and shape estimation. %

\vspace{1mm}
\noindent \textbf{Merge vs. Standard Convolutions for Shape Estimation} %
Another technical innovation in this work is the use of MergeConv blocks (\sect{sec:shape_estimation}) in the shape estimation network instead of standard convolution. %
Overall the depth estimation error decreased from a MSE of $0.021$ to $0.016$ and the normal MSE from $0.026$ to $0.021$. %

\subsection{Comparisons with state-of-the-art}
\vspace{-2mm}
As per our knowledge, we are the first work that uses two-shot images as input and does complete SVBRDF estimation, including specular color and shape estimation for objects. %
Most existing closely related techniques usually use a single flash image as input and either work only on flat surfaces~\cite{Deschaintre2018, Deschaintre2019, Li2018, Li2017}, or do not estimate the specular color~\cite{Li2018a}. %
Although our approach features a unique setting, we perform the comparisons with SIRFS~\cite{Barron2015}, Li \etal~\cite{Li2018a}, and RAFII~\cite{Nestmeyer2017} on SVBRDF and shape estimation. %
SIRFS~\cite{Barron2015} uses a no-flash single image as input and predicts diffuse albedo, shading, and shape using an optimization-based approach. %
RAFII~\cite{Nestmeyer2017} uses a single non-flash image to perform the intrinsic decomposition. Visual results are shown in the supplementary.
Based on a single flash image Li \etal~\cite{Li2018a} is a recent deep learning approach that predicts diffuse albedo, roughness, normal, and depth maps. 

Quantitative results on the 20 objects synthetic test dataset shown in \tbl{tbl:comparison} demonstrate the superior performance of our approach (Ours-CascadeNet) compared to both SIRFS~\cite{Barron2015} and Li \etal~\cite{Li2018a}. %
Since SIRFS predicts diffuse albedo only up to a scale factor, we also report scale-invariant MSE scores on diffuse albedo. %
\fig{fig:li_comparison_example} shows a visual comparison with Li \etal~\cite{Li2018a}. %
Our estimations are also visually closer to GT. %
Especially, we can observe clear visual differences in predicted diffuse albedos where the light information is separated much better in our result. %
Furthermore, the general shape of the object in the normal map of our method follows the contour of the croissant, while the method of Li \etal predicts a mostly flat shape. %
The details in the roughness and normal map, on the other hand, are not perfectly predicted by neither method. %

\fig{fig:barron_comparison} shows a visual comparison with SIRFS, where we again observe our method predictions to be closer to GT. %
Here, the improvements in the diffuse and normal map are apparent. %
The SIRFS method fails in this example to separate shape from shading. %

A visual comparison between Li \etal~\cite{Li2018a} on a real-world example from Yagiz \etal \cite{Yagiz2018} is shown in \fig{fig:real_world_comparison}. %
Our method seems to capture the object color as well as the shape better. %
The shape from Li \etal is predicted as a nearly flat surface. %
This is apparent in the novel re-rendering. Our predicted normal map is also smoother with fewer artifacts and follows the bottle shape closely. %

For evaluating depth prediction, we compare our depth estimates against those from a new state-of-the-art monocular depth network of MiDaS~\cite{Lasinger2019}. MiDaS is trained with
several existing depth datasets and is quiet robust to different scene types. MiDaS~\cite{Lasinger2019} predicts the relative depth, and for comparisons, a scale-shift invariant MSE metric is used. \tbl{tbl:comparison} shows the results indicating better depth estimations using our approach. We present qualitative results in the supplementary.

\vspace{1mm}
\noindent \textbf{Mobile capture and inference} %
To further showcase our real-world performance, \fig{fig:real_world_mobile} presents an example captured with our mobile application. %
As seen, most parameters are plausible. %
The lid on top of the electric kettle is, however, estimated slightly too far away in the depth map. %
This can be attributed to the 'deep is dark' ambiguity. %
Here, we want to point out that there is an additional challenge of an unknown mobile camera capture pipeline. %
A RAW image capture would avoid most of the unknown image pre-processing in modern cameras. %

%% file: figures/evaluation/syn_li_comparison/comparison.tex
\renewcommand{\arraystretch}{0.0}%
\begin{tabular}{@{}>{\centering\arraybackslash}m{4pt}>{\centering\arraybackslash}m{45pt}|>{\centering\arraybackslash}m{4pt}>{\centering\arraybackslash}m{45pt}>{\centering\arraybackslash}m{45pt}>{\centering\arraybackslash}m{45pt}>{\centering\arraybackslash}m{45pt}>{\centering\arraybackslash}m{45pt}>{\centering\arraybackslash}m{45pt}>{\centering\arraybackslash}m{45pt}@{}}%
\toprule%
&&&\scriptsize MSE: 0.029&&\scriptsize MSE: 0.031&\scriptsize MSE: 0.020&\scriptsize MSE: 0.012&\scriptsize MSE: 3.006& \\%
\rotatebox[origin=c]{90}{\scriptsize Flash} &\includegraphics[width=45pt]{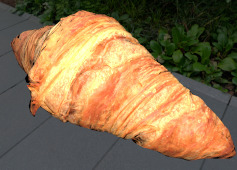} %
& \rotatebox[origin=c]{90}{\scriptsize Li et al.} %
& \includegraphics[width=45pt]{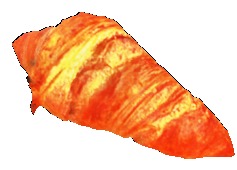} %
& \scriptsize Not estimated %
& \includegraphics[width=45pt]{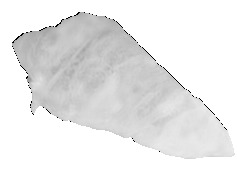} %
& \includegraphics[width=45pt]{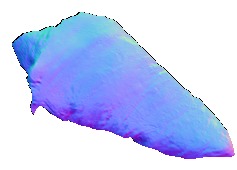} %
& \includegraphics[width=45pt]{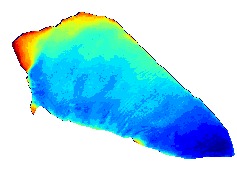} %
& \includegraphics[width=45pt]{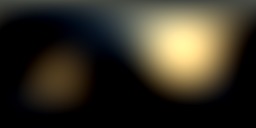} %
& \includemedia[ %
    width=45pt, %
    height=45pt, %
    playbutton=none, %
    activate=pagevisible, %
    deactivate=pageinvisible, %
    addresource=figures/evaluation/syn_li_comparison/li/814075_1_moving.mp4, %
    passcontext, %
    flashvars={source=figures/evaluation/syn_li_comparison/li/814075_1_moving.mp4&loop=true&autoPlay=true} %
  ]{\includegraphics[width=45pt,width=45pt]{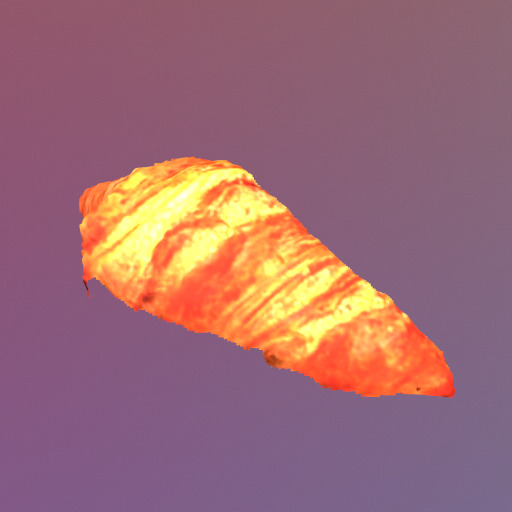}}{VPlayer.swf}\\%

\rotatebox[origin=c]{90}{\scriptsize No-Flash} & \includegraphics[width=45pt]{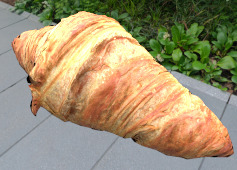} %
& \rotatebox[origin=c]{90}{\scriptsize GT} %
& \includegraphics[width=45pt]{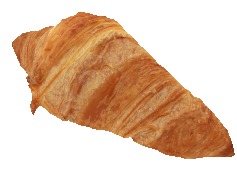} %
& \includegraphics[width=45pt]{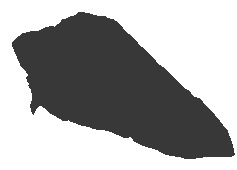} %
& \includegraphics[width=45pt]{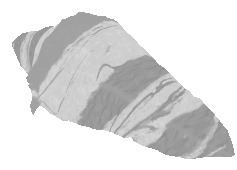} %
& \includegraphics[width=45pt]{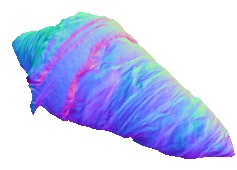} %
& \includegraphics[width=45pt]{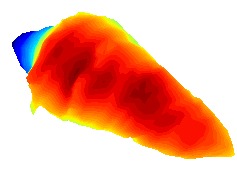} %
&\includegraphics[width=45pt]{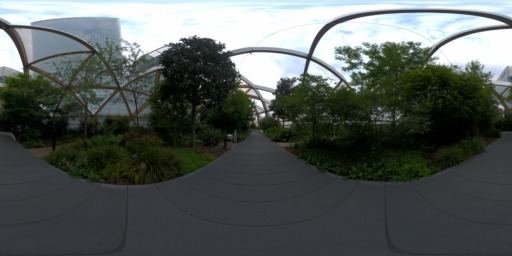} %
& \includemedia[ %
    width=45pt, %
    height=45pt, %
    playbutton=none, %
    activate=pagevisible, %
    deactivate=pageinvisible, %
    addresource=figures/evaluation/syn_li_comparison/gt/814075_1_moving.mp4, %
    passcontext, %
    flashvars={source=figures/evaluation/syn_li_comparison/gt/814075_1_moving.mp4&loop=true&autoPlay=true} %
  ]{\includegraphics[width=45pt,width=45pt]{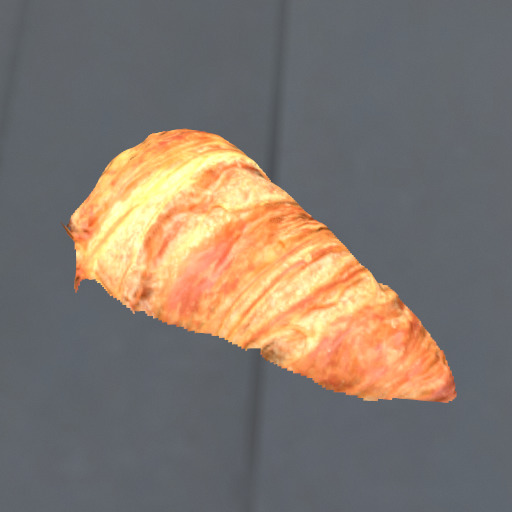}}{VPlayer.swf} \\ %
\rotatebox[origin=c]{90}{\scriptsize Mask} & \includegraphics[width=45pt]{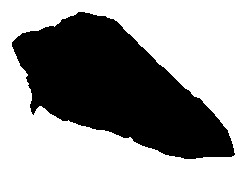} %
&\rotatebox[origin=c]{90}{\scriptsize Ours} %
& \includegraphics[width=45pt]{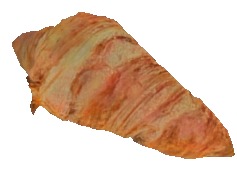} %
& \includegraphics[width=45pt]{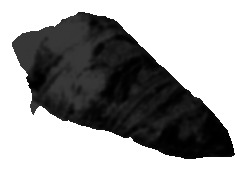} %
& \includegraphics[width=45pt]{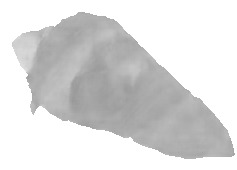} %
& \includegraphics[width=45pt]{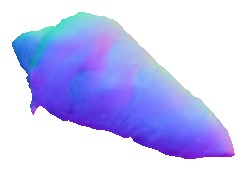} %
& \includegraphics[width=45pt]{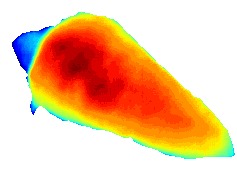} %
& \includegraphics[width=45pt]{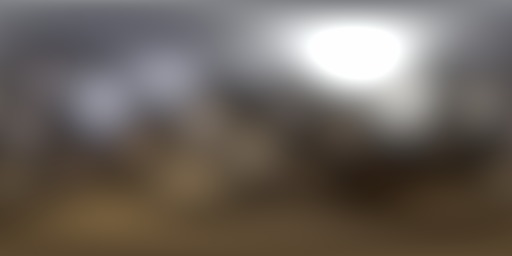} %
& \includemedia[ %
    width=45pt, %
    height=45pt, %
    playbutton=none, %
    activate=pagevisible, %
    deactivate=pageinvisible, %
    addresource=figures/evaluation/syn_li_comparison/own/814075_1_moving.mp4, %
    passcontext, %
    flashvars={source=figures/evaluation/syn_li_comparison/own/814075_1_moving.mp4&loop=true&autoPlay=true} %
  ]{\includegraphics[width=45pt,width=45pt]{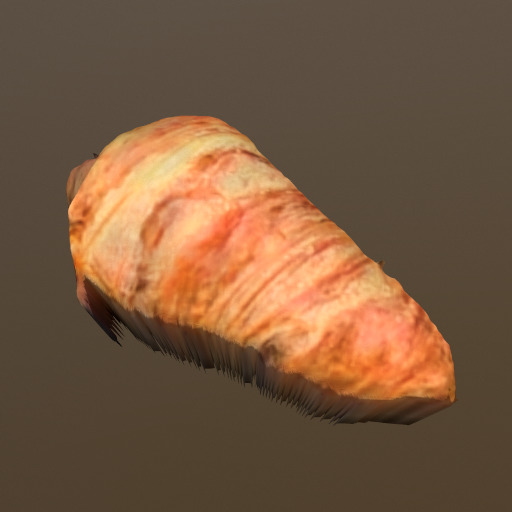}}{VPlayer.swf} \\ %
&&&\scriptsize MSE: 0.005&\scriptsize MSE: 0.020&\scriptsize MSE: 0.010&\scriptsize MSE: 0.011&\scriptsize MSE: 0.001&\scriptsize MSE: 2.900& \\
\midrule
& \scriptsize Input&&\scriptsize Diffuse&\scriptsize Specular&\scriptsize Roughness&\scriptsize Normal&\scriptsize Depth&\scriptsize Illumination&\scriptsize Re{-}render\\
\bottomrule%
\end{tabular}%

%% file: figures/evaluation/real_world_li_comparison/RealWorldParams.tex
\renewcommand{\arraystretch}{0.0}%
\begin{tabular}{@{}>{\centering\arraybackslash}m{2pt}>{\centering\arraybackslash}m{54pt}|>{\centering\arraybackslash}m{2pt}>{\centering\arraybackslash}m{54pt}>{\centering\arraybackslash}m{54pt}>{\centering\arraybackslash}m{54pt}>{\centering\arraybackslash}m{54pt}>{\centering\arraybackslash}m{54pt}>{\centering\arraybackslash}m{54pt}@{}}%
\toprule%
\rotatebox[origin=c]{90}{\scriptsize No-Flash}
& \includegraphics[height=54pt,width=54pt]{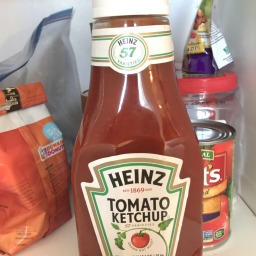}%
& \rotatebox[origin=c]{90}{\scriptsize Li et al.}
& \includegraphics[height=54pt]{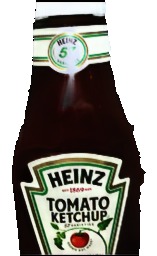} 
& \scriptsize Not estimated
& \includegraphics[height=54pt]{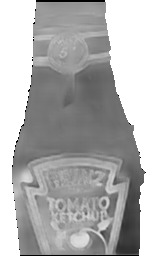} 
& \includegraphics[height=54pt]{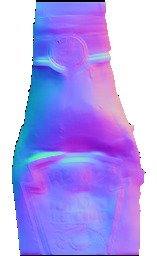} 
& \includegraphics[height=54pt]{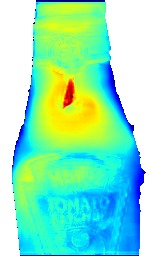}
& \includemedia[ %
width=54pt, %
height=54pt, %
playbutton=none, %
activate=pagevisible,%
deactivate=pageinvisible,%
addresource=figures/evaluation/real_world_li_comparison/li/Objects_545_moving.mp4, %
passcontext, %
flashvars={source=figures/evaluation/real_world_li_comparison/li/Objects_545_moving.mp4&loop=true&autoPlay=true} %
]{\includegraphics[width=54pt,width=54pt]{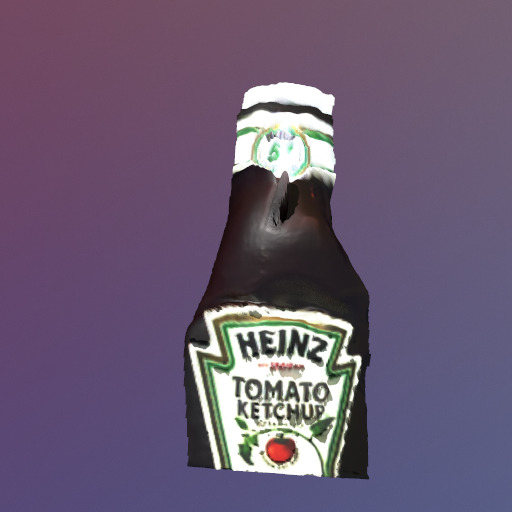}}{VPlayer.swf} \tabularnewline
\rotatebox[origin=c]{90}{\scriptsize Flash}
& \includegraphics[height=54pt,width=54pt]{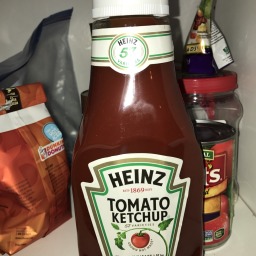}%
& \rotatebox[origin=c]{90}{\scriptsize Ours}
& \includegraphics[height=54pt]{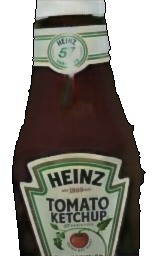} % Diffuse
& \includegraphics[height=54pt]{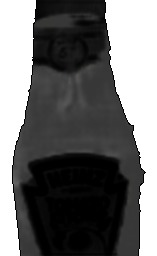} % Specular
& \includegraphics[height=54pt]{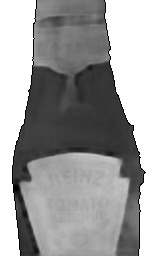} % Roughness
& \includegraphics[height=54pt]{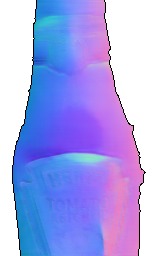} % Normal
& \includegraphics[height=54pt]{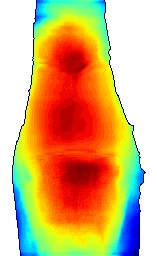}
& \includemedia[ %
width=54pt, %
height=54pt, %
playbutton=none, %
activate=pagevisible,%
deactivate=pageinvisible,%
addresource=figures/evaluation/real_world_li_comparison/own/Objects_545_moving.mp4, %
passcontext, %
flashvars={source=figures/evaluation/real_world_li_comparison/own/Objects_545_moving.mp4&loop=true&autoPlay=true} %
]{\includegraphics[width=54pt,width=54pt]{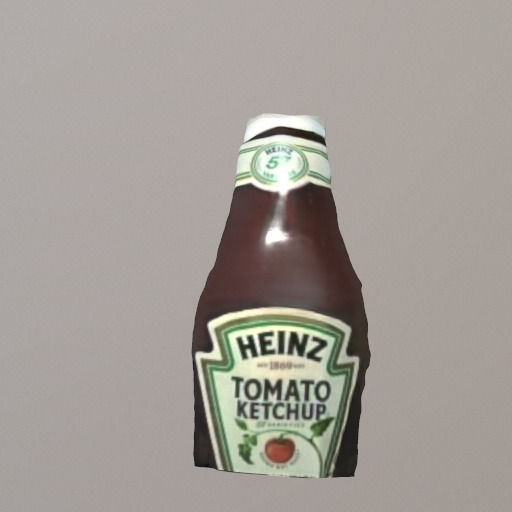}}{VPlayer.swf} \tabularnewline
\midrule
 & \scriptsize Input & &  \scriptsize Diffuse & \scriptsize Specular & \scriptsize Roughness & \scriptsize Normal & \scriptsize Depth & \scriptsize Novel Re-Render \\%
\bottomrule%
\end{tabular}%

%% file: figures/evaluation/real_world_own/RealMobileExample.tex
\setlength{\tabcolsep}{0pt}
\renewcommand{\arraystretch}{0}%
\begin{tabular}{@{}ccc@{\hskip 2pt}c@{\hskip 2pt}ccccc@{}}%
\toprule%
% Sample 1
\multirow{2}{*}[28pt]{\includegraphics[width=54pt, width=54pt]{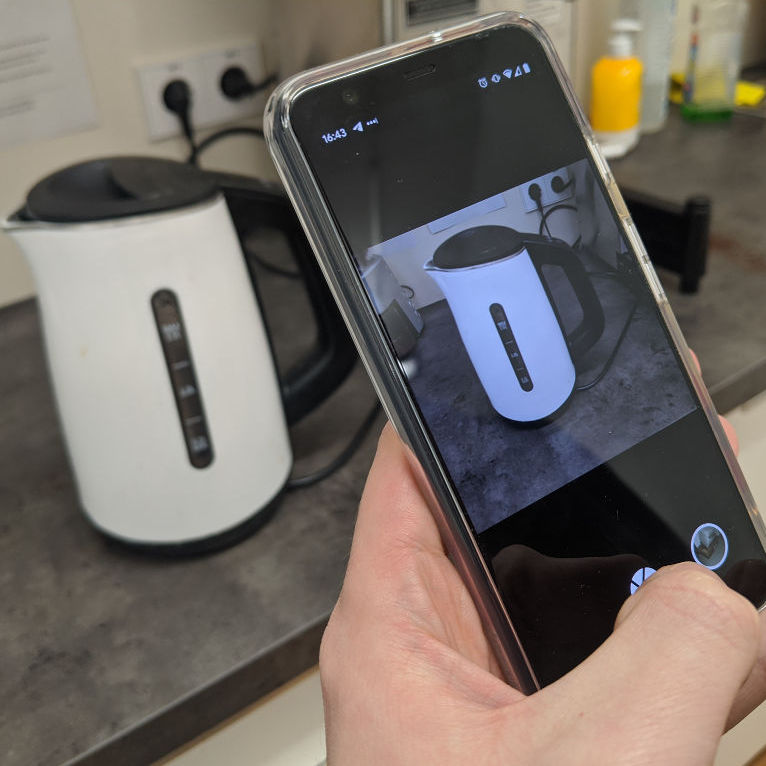}} %
& \multirow{2}{*}[28pt]{\includegraphics[height=54pt,width=54pt]{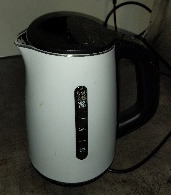}} %
& \includegraphics[height=28pt,width=28pt]{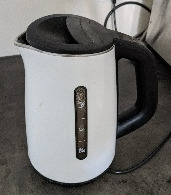} %
& \multirow{2}{*}[28pt]{\includegraphics[height=54pt,width=54pt]{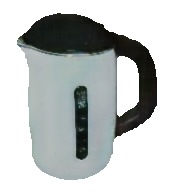}}%
& \multirow{2}{*}[28pt]{\includegraphics[height=54pt,width=54pt]{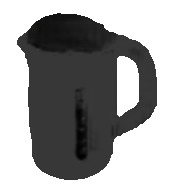}}%
& \multirow{2}{*}[28pt]{\includegraphics[height=54pt,width=54pt]{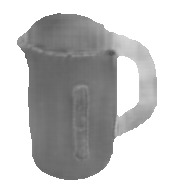}}%
& \multirow{2}{*}[28pt]{\includegraphics[height=54pt,width=54pt]{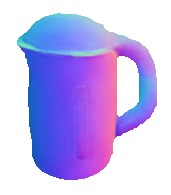}}%
& \multirow{2}{*}[28pt]{\includegraphics[height=54pt,width=54pt]{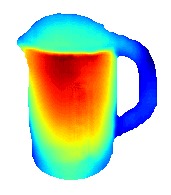}}
& \multirow{2}{*}[28pt]{%
\includemedia[ %
width=54pt, %
height=54pt, %
playbutton=none, %
activate=pagevisible, %
deactivate=pageinvisible, %
addresource=figures/evaluation/real_world_own/16_moving.mp4, %
passcontext, %
flashvars={source=figures/evaluation/real_world_own/16_moving.mp4&loop=true&autoPlay=true} %
]{\includegraphics[width=54pt,width=54pt]{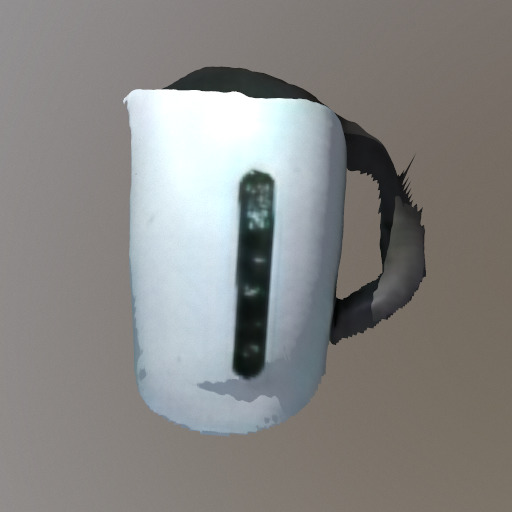}}{VPlayer.swf} %
} \tabularnewline
& & \includegraphics[height=28pt,width=28pt]{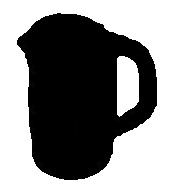} & \\
\midrule%
\scriptsize Capturing & \scriptsize Flash & \pbox{28pt}{\scriptsize No-Flash\\Mask} &  \scriptsize Diffuse & \scriptsize Specular & \scriptsize Roughness & \scriptsize Normal & \scriptsize Depth & \scriptsize Re-Render \\%
\bottomrule%
\end{tabular}%

%% file: sec_conclusion.tex
\vspace{-2mm}
\section{Conclusion}
\label{sec:conclusion}
\vspace{-2mm}

We proposed a novel cascaded network design coupled with guided prediction networks for SVBRDF and shape estimation from two-shot images. %
Our key insight is that the separation of tasks and stage-wise prediction can lead to significantly better results compared to joint estimation with a single large network. %
We use a two-shot capture setting, which is practical and helps in estimating higher quality SVBRDF and shape compared to existing works. %
All of our image capture, network inference, and rendering can be easily implemented on mobile hardware. %
Another key contribution is the creation of large-scale synthetic training data with domain-randomized geometry and carefully collected materials. %
We show that networks trained on this data can generalize well to real-world objects. %
In the future, we would like to tackle the SVBRDF estimation of more complex mirror-like objects by incorporating reflection removal techniques and anisotropic BRDF models. %

\vspace{-3mm}
\paragraph{Acknowledgement}
This work was partly funded by the Deutsche Forschungsgemeinschaft (German Research Foundation) - Projektnummer 276693517 - SFB 1233. We thank Ben Eckart for his help in the supplementary video.